\documentclass[runningheads]{llncs}
\usepackage{graphicx}
\usepackage{siunitx}
\usepackage{color}
\usepackage{amssymb}
\usepackage{amsmath, xparse}
\usepackage{makecell}
\usepackage{lipsum}
\usepackage{caption}
\usepackage{subcaption}
\usepackage{booktabs}
\usepackage{multirow}
\newcommand{\bfsection}[1]{\vspace*{0.1cm}\noindent\textbf{#1.}}
\newcommand{\red}[1]{\textcolor{red}{#1}}

\usepackage{xspace}
\makeatletter
\DeclareRobustCommand\onedot{\futurelet\@let@token\@onedot}
\def\@onedot{\ifx\@let@token.\else.\null\fi\xspace}

\def\eg{\emph{e.g}\onedot} 
\def\ie{\emph{i.e}\onedot}

\def\etal{\emph{et al}\onedot}
\makeatother

\usepackage[x11names]{xcolor}
\usepackage[pagebackref=true,breaklinks=true,colorlinks,bookmarks=false]{hyperref}
\hypersetup{
     colorlinks = true,
     linkcolor = red,
     anchorcolor = black,
     citecolor = SpringGreen4,
     filecolor = black,
     urlcolor = black
     }
\usepackage{circledsteps}

\newcommand*{\affaddr}[1]{#1} % No op here. Customize it for different styles.
\newcommand*{\affmark}[1][*]{\textsuperscript{#1}}

\makeatletter
\def\@fnsymbol#1{\ensuremath{\ifcase#1\or *\or \dagger\or  \ddagger\or
   \mathsection\or \mathparagraph\or \|\or **\or \dagger\dagger
   \or \ddagger\ddagger \else\@ctrerr\fi}}
\makeatother

\begin{document}
\title{Instance Segmentation of Unlabeled Modalities via Cyclic Segmentation GAN}
\titlerunning{CySGAN}
\author{
Leander Lauenburg\affmark[1,2]\thanks{These authors contributed equally to this work.}\and
Zudi Lin\affmark[1*]\thanks{Corresponding author. Email: \email{linzudi@g.harvard.edu}}\and
Ruihan Zhang\affmark[3]\and
Márcia dos Santos\affmark[4]\and
Siyu Huang\affmark[1]\and
Ignacio Arganda-Carreras\affmark[5,6,7]\and
Edward S. Boyden\affmark[3,8]\and
Hanspeter Pfister\affmark[1]\and
Donglai Wei\affmark[9]
}
\authorrunning{Lauenburg \etal}
\institute{
\affaddr{\affmark[1]Harvard University} \
\affaddr{\affmark[2]Technical University of Munich} \
\affaddr{\affmark[3]MIT} \
\affaddr{\affmark[4]University of Cambridge} \
\affaddr{\affmark[5]Donostia International Physics Center} \
\affaddr{\affmark[6]University of the Basque Country} \
\affaddr{\affmark[7]Ikerbasque, Basque Foundation for Science} \
\affaddr{\affmark[8]HHMI} \
\affaddr{\affmark[9]Boston College}
}
\maketitle              % typeset the header of the contribution
% \dw{"cyclic segmentation" is confusing. ppl may think that we apply cyclic loss to segmentation or the seg goes through a cycle. maybe "seg consistency loss for xx". Plus, the title doesn't differentiate us from SUSAN}
%
\begin{abstract}
    Instance segmentation for unlabeled imaging modalities is a challenging but essential task as collecting expert annotation can be expensive and time-consuming. Existing works segment a new modality by either deploying a pre-trained model optimized on diverse training data or conducting domain translation and image segmentation as two independent steps. In this work, we propose a novel {\em Cyclic Segmentation} Generative Adversarial Network (\textbf{CySGAN}) that conducts image translation and instance segmentation jointly using a unified framework. Besides the CycleGAN losses for image translation and supervised losses for the annotated source domain, we introduce additional self-supervised and segmentation-based adversarial objectives to improve the model performance by leveraging unlabeled target domain images. We benchmark our approach on the task of 3D {\em neuronal nuclei} segmentation with annotated electron microscopy (EM) images and unlabeled expansion microscopy (ExM) data. Our CySGAN outperforms both pre-trained generalist models and the baselines that sequentially conduct image translation and segmentation. Our implementation and the newly collected, densely annotated ExM nuclei dataset, named {\em NucExM}, are available at \url{https://connectomics-bazaar.github.io/proj/CySGAN/index.html}.
    \keywords{3D Instance Segmentation  \and  Unsupervised Domain Adaptation  \and Expansion Microscopy (ExM) \and Electron Microscopy (EM)}
\end{abstract}
% \and Image Translation
% Nucleus Segmentation \and
\section{Introduction}

3D Instance segmentation of cell nuclei is an essential topic attracting both biomedical and computer vision researchers~\cite{rivron2018blastocyst,caicedo2019nucleus,stringer2021cellpose,weigert2020star,lin2021nucmm}. Supervised deep learning with in-domain annotations (\eg, U-Net~\cite{ronneberger2015u,cciccek20163d}) has become the dominant methodology for common imaging modalities. However, for novel imaging techniques, \eg, expansion microscopy (ExM)~\cite{chen2015expansion}, such an approach is less applicable to newly collected large-scale data due to the high annotation cost.

\begin{figure}[t]
\centering
\includegraphics[width=\columnwidth]{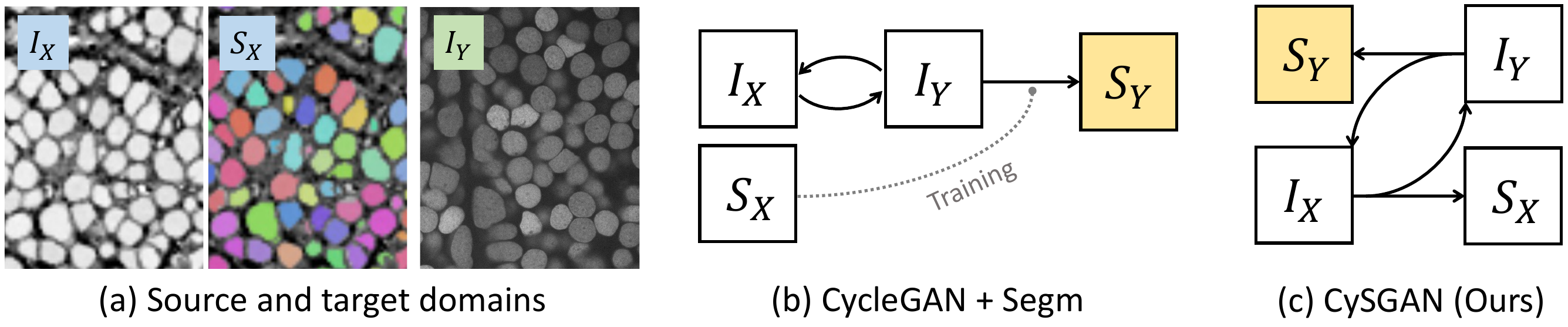}
\caption{
Overview of task and methods. {\bf (a)} We aim to segment an unlabeled target domain ($I_Y$) by leveraging the images ($I_X$) and masks ($S_X$) in the source domain. Instead of {\bf (b)} conducting image translation (\eg, via CycleGAN~\cite{zhu2017unpaired}) and instance segmentation as two separate steps, we propose {\bf (c)} the CySGAN framework to unify the two functionalities, optimized with both image translation as well as supervised and {\em semi-supervised} segmentation losses. 
% \dw{can we use one color for one model, so that ppl can see (b) needs 3 models and (c) only needs two model. (c) itself is not new, maybe add a dotted connection between $S_X$ and $S_Y$ for the proposed consistency loss?}
}\label{fig:teaser}
\end{figure}

To address the challenges, two common approaches aim to make use of existing labels for the unlabeled domain. One approach is to train a supervised model on diverse datasets and then directly apply it to the new domain~\cite{stringer2021cellpose,weigert2020star}. However, they can hardly be adapted to a new domain without label images. 
The other way makes use of unpaired image-to-image translation models like CycleGAN~\cite{zhu2017unpaired}. First, source images are translated to match the target domain distribution, aiming to be indistinguishable from the target domain while keeping the source structures. Next, the masks and translated images are used to train a supervised model and segment target images\footnote{The opposite way, which transfers the target images to the source domain and applies a supervised model trained on the source, is also reasonable. However, it is used less often as both the translation and segmentation models are required in inference.} (Fig.~\ref{fig:teaser}\red{b}).
However, the segmentation depends on a translation model optimized regardless of the downstream task, while two separate modules also increase the pipeline complexity. 

In this work, we propose a {\em Cyclic Segmentation} Generative Adversarial Network (CySGAN) that unifies image translation and segmentation to tackle unlabeled modalities (Fig.~\ref{fig:teaser}\red{c}). For both the source and target domains, we train a single 3D U-Net~\cite{cciccek20163d} that takes only images as input but outputs both segmentation and translated images simultaneously. Besides optimizing the translation and supervised segmentation losses as previous work~\cite{liu2019susan}, we introduce structural consistency and segmentation-based adversarial losses to better leverage the unlabeled domain images, connecting ideas from {\em semi-supervised} segmentation. Moreover, to incorporate data augmentations shown to enhance segmentation performance~\cite{lee2017superhuman,lin2021nucmm}, we enforce the cycle consistency~\cite{zhu2017unpaired} of the reconstructed images to the clean images instead of the augmented ones, which acts as regularization and enable the model to restore corrupted regions.

In addition, we curated and annotated two expansion microscopy (ExM) image volumes from an adult zebrafish brain tissue with dense cell nuclei ($I_Y$ in Fig.~\ref{fig:teaser}\red{a}). These two volumes are complemented by a publicly available and labeled electron microscopy (EM) dataset ($I_X$ and $S_X$ in Fig.~\ref{fig:teaser}\red{a}). Without any annotation for the ExM domain, our CySGAN outperforms models pretrained on diverse datasets and the methods that conduct translation and segmentation separately. We release our code and the new {\em NucExM} dataset for future research.

\subsection{Related Works}

\bfsection{Unpaired Image-to-Image Translation}
Paired images from different domains are expensive or even infeasible to obtain. Therefore, {\em unpaired} image-to-image translation~\cite{isola2017image,zhu2017unpaired} based on Generative Adversarial Networks (GAN)~\cite{goodfellow2014generative} becomes a sensible methodology to transfer source images to the target distribution. 
CycleGAN~\cite{zhu2017unpaired} achieves impressive performance by ensuring {\em cycle consistency} when transferring translated images back to the input domain. Further improvements including shared high-level layers~\cite{liu2017unsupervised} and latent space alignment~\cite{hoffman2018cycada}. We refer readers to the survey by Pang \etal~\cite{pang2021image} for a more detailed discussion on image-to-image translation. In particular, our work focus on combining image translation with segmentation models to tackle unlabeled modalities.

\bfsection{Instance Segmentation of 3D Microscopy}
3D instance segmentation from microscopy images is challenging due to the dense distribution of objects and unavoidable physical limitations in imaging (\eg, data is frequently anisotropic). Recent learning-based approaches tackle these challenges by first optimizing CNN-based models to predict representations calculated from the instance masks, including object boundary~\cite{ciresan2012deep,ronneberger2015u,wei2020mitoem}, affinity map~\cite{turaga2009maximin,lee2017superhuman}, star-convex distance~\cite{weigert2020star}, flow-field~\cite{stringer2021cellpose} and the combination of multiple representations~\cite{lin2021nucmm}. Watershed transform~\cite{cousty2008watershed,zlateski2015image} and graph partition~\cite{krasowski2017neuron} can then be applied to convert the predicted representations into instance masks. However, most existing works train the segmentation models in a supervised learning manner, which becomes infeasible considering the cost of acquiring expert annotations for new modalities. Our work focuses on unifying segmentation approaches with image translation to segment instances in new domains via unsupervised domain adaptation.

\bfsection{Combining Translation and Segmentation}
Segmenting unlabeled domains via image translation is a practical methodology. Chartsias \etal~\cite{chartsias2017adversarial} design a two-stage framework that first translates label images to the unlabeled domain using CycleGAN~\cite{zhu2017unpaired}, then trains a separate segmentation model using the synthesized images and original ground-truth label. CyCADA~\cite{hoffman2018cycada} and EssNet~\cite{huo2018adversarial} improve the sequential model by jointly optimizing the translation and segmentation networks. However, using two separate models increases the system complexity in training and deployment. The concept to simultaneously conduct translation and segmentation has been explored in SUSAN~\cite{liu2019susan}, but our work differs from it in two main aspects. First, SUSAN is for 2D semantic segmentation, while our work focuses on the more challenging 3D instance segmentation. Second, SUSAN only applies supervised segmentation losses to the annotated domain, while our CySGAN leverages structural consistency and segmentation-based adversarial losses for the unlabeled domain in the absence of ground-truth labels.
\section{Method}

Suppose we have an annotated {\em source} domain $X=(I_X, S_X)$ where $I_X$ and $S_X$ denote the images and paired segmentation labels, respectively. Then for an unlabeled {\em target} domain $Y$ with only $I_Y$, the goal is to generate the instance segmentation $S_Y$ without any manual annotations in the $Y$ domain. An idea is to first synthesize images $I_{Y^\prime} = F(I_X)$ that are {\em indistinguishable} from the distribution of $I_Y$ but keep the instance structure in $S_X$. Then a supervised model can be optimized using $(I_{Y^\prime}, S_X)$ pairs to predict $S_Y$ from $I_Y$ (Fig.~\ref{fig:teaser}\red{b}). 

Although it is straightforward to conduct translation and segmentation {\em sequentially}, the translation model is not designed with an end task in mind and can propagate errors to the second step. Besides, the two separate modules make the system complicated in training and deployment. Thus, we propose a framework to finish translation and instance segmentation {\em simultaneously} using two generators that output both translated images and segmentation (Fig.~\ref{fig:teaser}\red{c}):
\begin{equation}\label{eqn:joint}
    F: I_X \rightarrow (I_Y, S_X)\quad\quad\quad
    B: I_Y \rightarrow (I_X, S_Y)
\end{equation}
We denote the proposed framework as the {\em cyclic segmentation} GAN (CySGAN). Specifically, for an image ${x_i\sim I_X}$, we have $[\hat{y}_i, \hat{x}_s]=F(x_i)$, where $\hat{y}_i$ is the synthesized image, $\hat{x}_s$ contains the predicted instance representations (will elaborate later), and $[\hat{y}_i, \hat{x}_s]$ is their concatenation (a single model outputs them as different channels). For clarity in formulations, we also denote $\hat{y}_i=F(x_i)_{[\text{I}]}$ and $\hat{x}_s=F(x_i)_{[\text{S}]}$.
Note that $B(F(x_i))$ is no longer a valid expression as both models take only an image as input but output the translated image and segmentation. 

We optimize $F$ and $B$ together with the necessary discriminators and segment $I_Y$ with $B$. Our design largely simplifies the sequential framework with two isolated steps. Different from standard image translation, the two domains are {\em asymmetric}, as $X$ is labeled, while $Y$ is unlabeled. We thus apply similar image translation losses but unique segmentation losses for $X$ and $Y$ domains. 

\subsection{Image Translation Losses}\label{sec:trans_loss}
We denote $F$ as the {\em forward} generator. Since paired $I_X$ and $I_Y$ are difficult or even infeasible to obtain, $F$ is usually optimized using the {\em adversarial} loss:
\begin{equation}\label{eqn:gan}
    \mathcal{L}_{GAN}(F, D_Y^I) =
    \log D_Y^I(y_i) +
    \log(1-D_Y^I(\hat{y}_i)),\quad
    \hat{y}_i=F(x_i)_{[\text{I}]}
\end{equation}
where $D_{Y}^{I}$ is the $I_Y$ discriminator, while $y_i$ and $\hat{y}_i$ are true and synthesized images, respectively. Following CycleGAN~\cite{zhu2017unpaired}, we also use the {\em backward} generator $B$ and an $I_X$ discriminator $D_{X}^{I}$ to symmetrically optimize $\mathcal{L}_{GAN}(B, D_X^I)$, as well as enforcing the {\em cycle-consistency} loss for the images in both domains:
\begin{equation}\label{eqn:cyclic}
    \mathcal{L}_{cyc}(F, B) = \|B(\hat{y}_i)_{[\text{I}]}-x_i\|_1 + \|F(\hat{x}_i)_{[\text{I}]}-y_i\|_1
\end{equation}
The losses enable the models to transfer images between $I_X$ and $I_Y$ distributions.

\subsection{Instance Segmentation Losses}\label{sec:seg_loss}

\begin{figure}[t]
\centering
\includegraphics[width=\columnwidth]{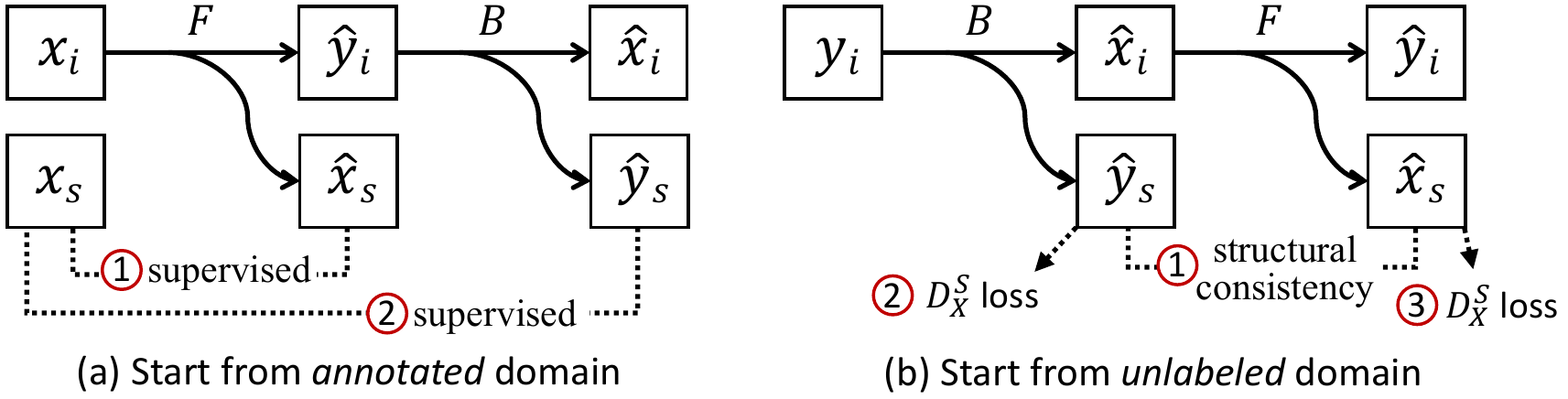}
\caption{
Segmentation losses. {\bf (a)} For an annotated image in $X$, we compute the supervised losses of predicted representations against the label. {\bf (b)} For an unlabeled image in $Y$, we enforce {\em structural consistency} between predicted representations (as the underlying structures should be shared) and also adversarial losses to improve the quality of predictions in the absence of paired labels.
}\label{fig:losses}
\end{figure}

\bfsection{Labeled Source Domain}
Instance segmentation approaches for microscopy images~\cite{stringer2021cellpose,weigert2020star,wei2020mitoem,lin2021nucmm} usually predict instance representations computed from the permutation-invariant labels and then apply a decoding algorithm to yield the masks. In this work, we follow U3D-BCD~\cite{lin2021nucmm} that predicts the {\em binary foreground mask} (B), {\em instance contour map} (C), and {\em signed distance transform} (D) as three output channels using a 3D U-Net~\cite{cciccek20163d}, which are decoded by a marker-controlled watershed (MW) algorithm. The B and C channels are optimized with the binary cross-entropy loss (BCE), while D is regressed with the mean squared error (MSE). Given an image-label pair $(x_i, x_s)$ sampled from $(I_X, S_X)$, the loss is
\begin{equation}\label{eqn:seg}
    \mathcal{L}_{seg}(F) = 
    \mathcal{L}_{bce}\left(F(x_i)^B_{[\text{S}]}, x_s^B\right) +
    \mathcal{L}_{bce}\left(F(x_i)^C_{[\text{S}]}, x_s^C\right) +
    \|F(x_i)^D_{[\text{S}]} - x_s^D\|_2^2
\end{equation}
where $x_s=[x_s^B, x_s^C, x_s^D]$ is the concatenation of the three representations. $\mathcal{L}_{seg}(F)$ and another segmentation loss $\mathcal{L}_{seg}(B)$ (based on the synthesized $\hat{y}_i$) are optimized by directly comparing $\hat{x}_s$ and $\hat{y}_s$ with $x_s$ from $S_X$ ({\small \Circled{1}} and {\small \Circled{2}} in Fig.~\ref{fig:losses}\red{a}).

The loss $\mathcal{L}_{seg}(B)$ effectively trains $B$ in a supervised manner to predict the segmentation representations. Moreover, this design is not restricted to a particular representation and can be easily modified to incorporate other approaches. % We then present a set of novel losses introduced to better leverage the {\em unlabeled} domain $Y$.

\bfsection{Unlabeled Target Domain}
Since $Y$ is unlabeled, it is impossible to apply the supervised losses for $X$. To further improve segmentation quality, we introduce a {\em structural consistency} loss between the segmentation outputs of both generators, $\hat{y}_s$ and $\hat{x}_s$ ({\small \Circled{1}} Fig.~\ref{fig:losses}\red{b}), as they should share identical underlying structures even if the inputs are from two modalities. This loss $\mathcal{L}_{sc}(F,B)$ is formulated as
\begin{equation}\label{eqn:struct_const}
    \mathcal{L}_{sc}(F,B) =
    \|B(y_i)_{[\text{S}]} - 
    F(B(y_i)_{[\text{I}]})_{[\text{S}]}\|_1
\end{equation}
We also add structure-based adversarial losses ({\small \Circled{2}} and {\small \Circled{3}} in Fig.~\ref{fig:losses}\red{b}) to the predictions to enforce their distributional similarity with $S_X$ (called $\mathcal{L}_{GAN}(B, D_X^S)$ and $\mathcal{L}_{GAN}(F, D_X^S)$). Specifically, the discriminator $D_X^S$ takes the concatenation of all three representations to emphasize the correlation between them as the representations are calculated from the same instance masks. This also avoids using three independent discriminators that increase the system complexity. Those losses provided additional supervision in the absence of paired labels for $I_Y$.

Our method is connected to {\em semi-supervised} learning as we incorporate unlabeled images in optimization using losses without paired labels. We can also choose other semi-supervised objectives, \eg, augmentation consistency~\cite{sohn2020fixmatch}. Our work emphasizes the concept of leveraging unlabeled images in a unified translation-segmentation framework, while the specific design choices can vary.

\subsection{Implementation}
\begin{figure}[t]
\centering
\includegraphics[width=\columnwidth]{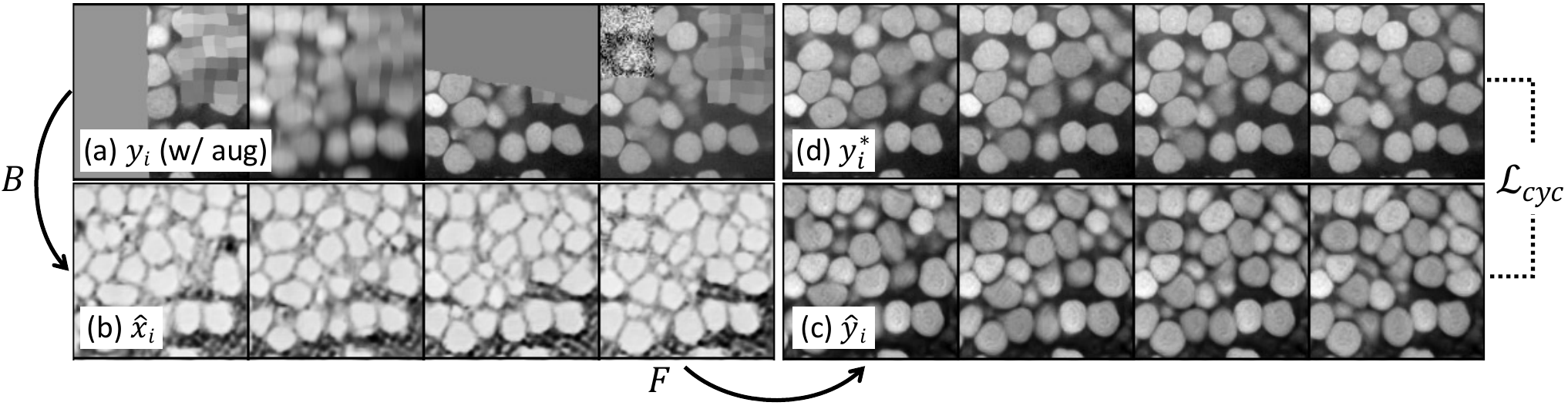}
\caption{
Training augmentations. We show four consecutive slices of {\bf (a)} augmented real $I_Y$ input, {\bf (b)} synthesized $I_X$ volume, {\bf (c)} reconstructed $I_Y$ volume and {\bf (d)} real $I_Y$ volume w/o augmentations. By forcing the cycle consistency of (c) to (d), the model learns to restore corrupted regions with 3D context.
}\label{fig:restore}
\end{figure}
The full objective of CySGAN is the sum of losses in Sec.~\ref{sec:trans_loss} and \ref{sec:seg_loss}, which is 
\begin{equation}\label{eqn:full_loss}
\begin{split}
    \mathcal{L} &= \underbrace{\mathcal{L}_{GAN}(F, D_Y^I) + \mathcal{L}_{GAN}(B, D_X^I) + \mathcal{L}_{cyc}(F, B)}_\text{image-to-image translation} + \underbrace{\mathcal{L}_{seg}(F) + \mathcal{L}_{seg}(B)}_\text{supervised segm}\\
    &+ \underbrace{\mathcal{L}_{sc}(F,B) + \mathcal{L}_{GAN}(B, D_X^S) + \mathcal{L}_{GAN}(F, D_X^S)}_\text{semi-supervised segm}
\end{split}
\end{equation}
We assign a uniform weight for all losses without tweaking. We use a 3D U-Net~\cite{cciccek20163d} for $F$ and $B$ (they have identical architectures, but the parameters are not shared). We use 3D convolutional discriminators, where $D_X$ and $D_Y$ have a single input channel for the gray-scale images, while $D_S$ has three input channels for the BCD representations. Following the CycleGAN~\cite{zhu2017unpaired} official code, we optimize the LSGAN~\cite{mao2017least} loss instead of the BCE GAN loss (Eqn.~\ref{eqn:gan}) for training stability. When calculating the segmentation losses, we detach the synthesized image to avoid the segmentation objectives affecting the image translation results. 

U3D-BCD~\cite{lin2021nucmm} uses multiple training augmentations like random missing, blurry and noisy regions (Fig.~\ref{fig:restore}\red{a}). We keep them for better segmentation quality. However, enforcing the cycle consistency (Eqn.~\ref{eqn:cyclic}) using corrupted images can cause ambiguity in translation. Therefore we stream the training images in both augmented and clean forms. As shown in Fig.~\ref{fig:restore} (each subfigure shows consecutive slices of a 3D volume), $B$ transfers augmented $y_i$ to $\hat{x}_i$, and $F$ reconstructs $\hat{x}_i$ to $\hat{y}_i$. Instead of calculating $\mathcal{L}_{cyc}(F, B)$ of $\hat{y}_i$ to $y_i$, we enforce its similarity to the clean $y_i^*$ (Fig.~\ref{fig:restore}\red{d}). Our models keep the augmentations for better segmentation and additionally learn to restore corrupted regions using 3D context. Our implementation of the proposed CySGAN framework is based on the {\em PyTorch Connectomics}~\cite{lin2021pytorch} open-source package. 

\section{Datasets}
\begin{table}[t]
\caption{\textbf{NucExM dataset.} We curated and densely annotated a {\em neuronal nuclei} segmentation dataset with two ExM volumes of zebrafish.}\label{tab:dataset}

\centering
\resizebox{\textwidth}{!}{% <------ Don't forget this %
\begin{tabular}{lccccc}
\toprule
Sample & ~\#Volumes~ & ~Volume Size (each)~ & ~Resolution ($\mu$m)~ & ~Ex. Ratio~ & ~\#Instances\\
\midrule
Zebrafish Brain & 2 & 2048$\times$2048$\times$255 & 0.325$\times$0.325$\times$2.5 & 7.0 & 9.6K+8.8K\\
\bottomrule
\end{tabular}
}
\end{table}

\bfsection{NucExM Dataset (Target)} We curated the saturated nuclei segmentation annotation for two expansion microscopy (ExM)~\cite{chen2015expansion} volumes by two experts from a day 7 post-fertilization (dpf) zebrafish brain, imaged with confocal microscopy. These volumes have an anisotropic resolution of $0.325\times 0.325\times 2.5\ \mu m$ in $(x,y,z)$ order, with an approximate tissue expansion factor of $7.0$. Thus the effective resolution becomes $0.046\times 0.046\times 0.357\ \mu m$. The two volumes are of size 2048$\times$2048$\times$255 voxels with 9.6K and 8.8K nuclei, respectively (Table~\ref{tab:dataset}).

\bfsection{Source Dataset and Resolution Matching} We use the NucMM-Z electron microscopy (EM) volume from the NucMM dataset~\cite{lin2021nucmm} as the source ($I_X$ and $S_X$ in Fig.~\ref{fig:teaser}\red{a}). The original NucMM-Z covers nearly a whole zebrafish brain at a resolution of $0.48\times 0.48\times 0.48\ \mu m$. Considering the different resolutions of two datasets, we crop a $200 \times 200 \times 255$ subvolume from NucMM-Z and upsample it to $512 \times 512 \times 255$, which contains 12K nuclei instances. During model training and inference, we downsample NucExM by $\times4$ along $x$ and $y$ axes to $512 \times 512 \times 255$, so that both the resolution and size match for the two datasets.

\bfsection{Evaluation Metric} 
Following common practice in instance segmentation~\cite{cordts2016cityscapes,lin2014microsoft}, we choose average precision (AP) as the evaluation metric. Specifically, for our 3D volumetric data, we choose AP-50 (\ie, AP with an IoU threshold of 0.5) and use the existing public implementation with improved efficiency~\cite{wei2020mitoem}. 
\section{Experiments}
\begin{table}[t]
\caption{
\textbf{Benchmark results on the NucExM dataset.} We compare both pretrained segmentation networks and translation-segmentation models using the AP scores. In the two-step approaches, we use  U3D-BCD~\cite{lin2021nucmm} for segmentation. {\bf Bold} and \underline{underlined} numbers denote the 1st and 2nd results.
}\label{tab:exp_main}

\centering
\resizebox{\textwidth}{!}{%
\begin{tabular}{lccccccc}
\toprule
\multirow{2}[2]{*}{Method}
& \multirow{2}[2]{*}{Cellpose}
& \multirow{2}[2]{*}{StarDist}
& \multicolumn{2}{c}{Histogram + Segm}
& \multicolumn{2}{c}{CycleGAN + Segm}
& \multirow{2}[2]{*}{\begin{tabular}{@{}c@{}}CySGAN \\ ({\bf Ours})\end{tabular}}
\\\cmidrule(lr){4-5} \cmidrule(lr){6-7}
& & 
& ~$I_X\rightarrow I_Y$~
& ~$I_Y\rightarrow I_X$~
& ~$I_X\rightarrow I_Y$~
& ~$I_Y\rightarrow I_X$~
\\

\midrule

AP-50 ($V_1$) & 0.644 & 0.816 & 0.807 & 0.804 & \underline{0.867} & 0.772 & {\bf 0.927} \\
AP-50 ($V_2$) & 0.765 & 0.875 & 0.826 & 0.816 & \underline{0.881} & 0.777 & {\bf 0.934} \\
\midrule
Average & 0.705 & 0.846 & 0.817 & 0.810 & \underline{0.874} & 0.775
 & {\bf 0.931} \\
\bottomrule
\end{tabular}
} % <-resize box
\end{table}

\bfsection{Methods in Comparison} We compare with Cellpose~\cite{stringer2021cellpose} and StarDist~\cite{weigert2020star} pretrained models using their official implementation. For methods that conduct translation and segmentation sequentially, we test both histogram matching and CycleGAN~\cite{zhu2017unpaired} as translation models. We use U3D-BCD~\cite{lin2021nucmm} for segmentation, which is consistent with CySGAN generators but without the output channel for translated images. Specifically, we test the $I_X\rightarrow I_Y$ version that transfers $I_X$ to $I_{Y^\prime}$ and trains a model in the target domain using synthesized images, and $I_Y\rightarrow I_X$ that transfers $I_Y$ to $I_{X^\prime}$ and predicts the segmentation using a model trained in the source. For CycleGAN and CySGAN, we only use one ExM volume ($V_1$) to optimize the models and directly run inference on the other volume ($V_2$) that is not used for optimization.

\begin{figure}[t]
\centering
\includegraphics[width=\columnwidth]{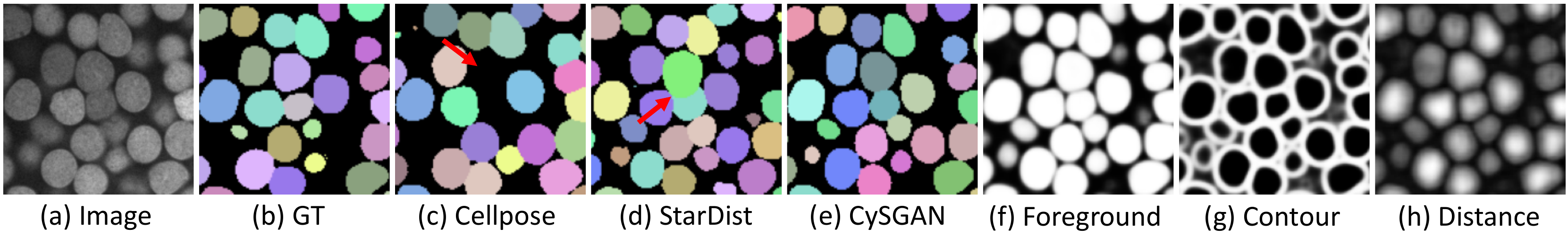}
\caption{
Visual comparisons. \textbf{(a)} ExM image, \textbf{(b)} ground-truth instances, \textbf{(c)} Cellpose~\cite{stringer2021cellpose}, \textbf{(d)} StarDist~\cite{weigert2020star} and \textbf{(e)} CySGAN results. We also show \textbf{(f-h)} predicted segmentation representations of U3D-BCD used with CySGAN.
}\label{fig:qual}
\end{figure}

% before converting them into masks

\bfsection{Results} Table~\ref{tab:exp_main} shows that our CySGAN outperforms both pretrained models and the sequential translation-segmentation models with either histogram matching or CycleGAN for translation. Specifically, CySGAN outperforms the second-best model (CycleGAN+Segm, $I_X\rightarrow I_Y$) by absolutely $5.7\%$, demonstrating the effectiveness of our proposed framework. The visual results in Fig.~\ref{fig:qual} show that Cellpose segmentation has obvious false negatives, while StarDist masks do not align well with instance boundaries and overlap with each other. We empirically find that the strong star-convex shape prior often overlooks other features like boundaries. The results also show that $I_X\rightarrow I_Y$ versions generally perform better than $I_Y\rightarrow I_X$ ones in sequential models.

\begin{table}[t]
\caption{
\textbf{Ablation study of CySGAN.} The results show the importance of data augmentations and semi-supervised segmentation losses in our CySGAN.
}\label{tab:exp_ablation}

\centering
\resizebox{1.0\textwidth}{!}{%
\begin{tabular}{lccc}
\toprule
Configuration~
& ~CySGAN w/o Augment~
& ~CySGAN w/o Semi-sup~
& ~CySGAN ({\bf Ours})~
\\
\midrule
AP-50 ($V_1$)
& 0.761 (\textbf{\textcolor{red!60}{-0.166}}) 
& 0.878 (\textbf{\textcolor{red!60}{-0.049}})
& {\bf 0.927}
\\
\bottomrule
\end{tabular}
} % <-resize box
\end{table}

\bfsection{Ablation Studies} We further validate two important design choices of CySGAN, including the semi-supervised segmentation losses for the {\em unlabeled} domain (Eqn.~\ref{eqn:full_loss}) and data augmentations (Fig.~\ref{fig:restore}). Table~\ref{tab:exp_ablation} shows that on the $V_1$ NucExM volume on which the models are optimized, removing either the training augmentations or the semi-supervised losses can result in obvious performance degradation of CySGAN, demonstrating the essentiality of those components.
\section{Conclusion}
In this work, we present CySGAN, a unified translation-segmentation framework optimized with image translation losses as well as supervised and semi-supervised instance segmentation losses to tackle unlabeled domains. Important future directions include segmenting target domains where the instance structures are significantly different from those from the source domain.

\subsection*{Acknowledgement}
This work has been partially supported by NSF awards IIS-1835231 and IIS-2124179 and NIH grant 5U54CA225088-03. Leander Lauenburg acknowledges the support from a fellowship within the IFI program of the German Academic Exchange Service (DAAD). Ignacio Arganda-Carreras acknowledges the support of the Beca Leonardo a Investigadores y Creadores Culturales 2020 de la Fundación BBVA. Edward S. Boyden acknowledges NIH 1R01EB024261, Lisa Yang, John Doerr, NIH 1R01MH123403, NIH 1R01MH123977, Schmidt Futures.

% reference
\bibliographystyle{splncs04.bst}
\bibliography{egbib.bib}

\begin{thebibliography}{10}
\providecommand{\url}[1]{\texttt{#1}}
\providecommand{\urlprefix}{URL }
\providecommand{\doi}[1]{https://doi.org/#1}

\bibitem{caicedo2019nucleus}
Caicedo, J.C., Goodman, A., Karhohs, K.W., Cimini, B.A., Ackerman, J.,
  Haghighi, M., Heng, C., Becker, T., Doan, M., McQuin, C., et~al.: Nucleus
  segmentation across imaging experiments: the 2018 data science bowl. Nature
  methods  \textbf{16}(12),  1247--1253 (2019)

\bibitem{chartsias2017adversarial}
Chartsias, A., Joyce, T., Dharmakumar, R., Tsaftaris, S.A.: Adversarial image
  synthesis for unpaired multi-modal cardiac data. In: International workshop
  on simulation and synthesis in medical imaging. pp. 3--13. Springer (2017)

\bibitem{chen2015expansion}
Chen, F., Tillberg, P.W., Boyden, E.S.: Expansion microscopy. Science
  \textbf{347}(6221),  543--548 (2015)

\bibitem{cciccek20163d}
{\c{C}}i{\c{c}}ek, {\"O}., Abdulkadir, A., Lienkamp, S.S., Brox, T.,
  Ronneberger, O.: 3d u-net: learning dense volumetric segmentation from sparse
  annotation. In: MICCAI. pp. 424--432. Springer (2016)

\bibitem{ciresan2012deep}
Ciresan, D., Giusti, A., Gambardella, L.M., Schmidhuber, J.: Deep neural
  networks segment neuronal membranes in electron microscopy images. In:
  NeurIPS. pp. 2843--2851 (2012)

\bibitem{cordts2016cityscapes}
Cordts, M., Omran, M., Ramos, S., Rehfeld, T., Enzweiler, M., Benenson, R.,
  Franke, U., Roth, S., Schiele, B.: The cityscapes dataset for semantic urban
  scene understanding. In: Proceedings of the IEEE conference on computer
  vision and pattern recognition. pp. 3213--3223 (2016)

\bibitem{cousty2008watershed}
Cousty, J., Bertrand, G., Najman, L., Couprie, M.: Watershed cuts: Minimum
  spanning forests and the drop of water principle. TPAMI  \textbf{31},
  1362--1374 (2008)

\bibitem{goodfellow2014generative}
Goodfellow, I., Pouget-Abadie, J., Mirza, M., Xu, B., Warde-Farley, D., Ozair,
  S., Courville, A., Bengio, Y.: Generative adversarial nets. Advances in
  neural information processing systems  \textbf{27} (2014)

\bibitem{hoffman2018cycada}
Hoffman, J., Tzeng, E., Park, T., Zhu, J.Y., Isola, P., Saenko, K., Efros, A.,
  Darrell, T.: Cycada: Cycle-consistent adversarial domain adaptation. In:
  International conference on machine learning. pp. 1989--1998. PMLR (2018)

\bibitem{huo2018adversarial}
Huo, Y., Xu, Z., Bao, S., Assad, A., Abramson, R.G., Landman, B.A.: Adversarial
  synthesis learning enables segmentation without target modality ground truth.
  In: 2018 IEEE 15th international symposium on biomedical imaging (ISBI 2018).
  pp. 1217--1220. IEEE (2018)

\bibitem{isola2017image}
Isola, P., Zhu, J.Y., Zhou, T., Efros, A.A.: Image-to-image translation with
  conditional adversarial networks. In: Proceedings of the IEEE conference on
  computer vision and pattern recognition. pp. 1125--1134 (2017)

\bibitem{krasowski2017neuron}
Krasowski, N., Beier, T., Knott, G., K{\"o}the, U., Hamprecht, F.A., Kreshuk,
  A.: Neuron segmentation with high-level biological priors. TMI
  \textbf{37}(4) (2017)

\bibitem{lee2017superhuman}
Lee, K., Zung, J., Li, P., Jain, V., Seung, H.S.: Superhuman accuracy on the
  snemi3d connectomics challenge. arXiv:1706.00120  (2017)

\bibitem{lin2014microsoft}
Lin, T.Y., Maire, M., Belongie, S., Hays, J., Perona, P., Ramanan, D.,
  Doll{\'a}r, P., Zitnick, C.L.: Microsoft coco: Common objects in context. In:
  European conference on computer vision. pp. 740--755. Springer (2014)

\bibitem{lin2021pytorch}
Lin, Z., Wei, D., Lichtman, J., Pfister, H.: Pytorch connectomics: A scalable
  and flexible segmentation framework for em connectomics. arXiv preprint
  arXiv:2112.05754  (2021)

\bibitem{lin2021nucmm}
Lin, Z., Wei, D., Petkova, M.D., Wu, Y., Ahmed, Z., Zou, S., Wendt, N.,
  Boulanger-Weill, J., Wang, X., Dhanyasi, N., et~al.: Nucmm dataset: 3d
  neuronal nuclei instance segmentation at sub-cubic millimeter scale. In:
  International Conference on Medical Image Computing and Computer-Assisted
  Intervention. pp. 164--174. Springer (2021)

\bibitem{liu2019susan}
Liu, F.: Susan: segment unannotated image structure using adversarial network.
  Magnetic resonance in medicine  \textbf{81}(5),  3330--3345 (2019)

\bibitem{liu2017unsupervised}
Liu, M.Y., Breuel, T., Kautz, J.: Unsupervised image-to-image translation
  networks. Advances in neural information processing systems  \textbf{30}
  (2017)

\bibitem{mao2017least}
Mao, X., Li, Q., Xie, H., Lau, R.Y., Wang, Z., Paul~Smolley, S.: Least squares
  generative adversarial networks. In: Proceedings of the IEEE international
  conference on computer vision. pp. 2794--2802 (2017)

\bibitem{pang2021image}
Pang, Y., Lin, J., Qin, T., Chen, Z.: Image-to-image translation: Methods and
  applications. IEEE Transactions on Multimedia  (2021)

\bibitem{rivron2018blastocyst}
Rivron, N.C., Frias-Aldeguer, J., Vrij, E.J., Boisset, J.C., Korving, J.,
  Vivi{\'e}, J., Truckenm{\"u}ller, R.K., Van~Oudenaarden, A.,
  Van~Blitterswijk, C.A., Geijsen, N.: Blastocyst-like structures generated
  solely from stem cells. Nature  \textbf{557}(7703),  106--111 (2018)

\bibitem{ronneberger2015u}
Ronneberger, O., Fischer, P., Brox, T.: U-net: Convolutional networks for
  biomedical image segmentation. In: MICCAI. pp. 234--241. Springer (2015)

\bibitem{sohn2020fixmatch}
Sohn, K., Berthelot, D., Carlini, N., Zhang, Z., Zhang, H., Raffel, C.A.,
  Cubuk, E.D., Kurakin, A., Li, C.L.: Fixmatch: Simplifying semi-supervised
  learning with consistency and confidence. Advances in Neural Information
  Processing Systems  \textbf{33},  596--608 (2020)

\bibitem{stringer2021cellpose}
Stringer, C., Wang, T., Michaelos, M., Pachitariu, M.: Cellpose: a generalist
  algorithm for cellular segmentation. Nature Methods  \textbf{18}(1),
  100--106 (2021)

\bibitem{turaga2009maximin}
Turaga, S.C., Briggman, K.L., Helmstaedter, M., Denk, W., Seung, H.S.: Maximin
  affinity learning of image segmentation. In: NeurIPS. pp. 1865--1873 (2009)

\bibitem{wei2020mitoem}
Wei, D., Lin, Z., Franco-Barranco, D., Wendt, N., et~al.: Mitoem dataset:
  Large-scale 3d mitochondria instance segmentation from em images. In:
  International Conference on Medical Image Computing and Computer-Assisted
  Intervention. pp. 66--76. Springer (2020)

\bibitem{weigert2020star}
Weigert, M., Schmidt, U., Haase, R., Sugawara, K., Myers, G.: Star-convex
  polyhedra for 3d object detection and segmentation in microscopy. In:
  Proceedings of the IEEE/CVF Winter Conference on Applications of Computer
  Vision. pp. 3666--3673 (2020)

\bibitem{zhu2017unpaired}
Zhu, J.Y., Park, T., Isola, P., Efros, A.A.: Unpaired image-to-image
  translation using cycle-consistent adversarial networks. In: Proceedings of
  the IEEE international conference on computer vision. pp. 2223--2232 (2017)

\bibitem{zlateski2015image}
Zlateski, A., Seung, H.S.: Image segmentation by size-dependent single linkage
  clustering of a watershed basin graph. arXiv:1505.00249  (2015)

\end{thebibliography}

\appendix
\renewcommand{\thefigure}{A-\arabic{figure}}
\setcounter{figure}{0}

\newpage
\begin{figure}[t]
\centering
\includegraphics[width=\columnwidth]{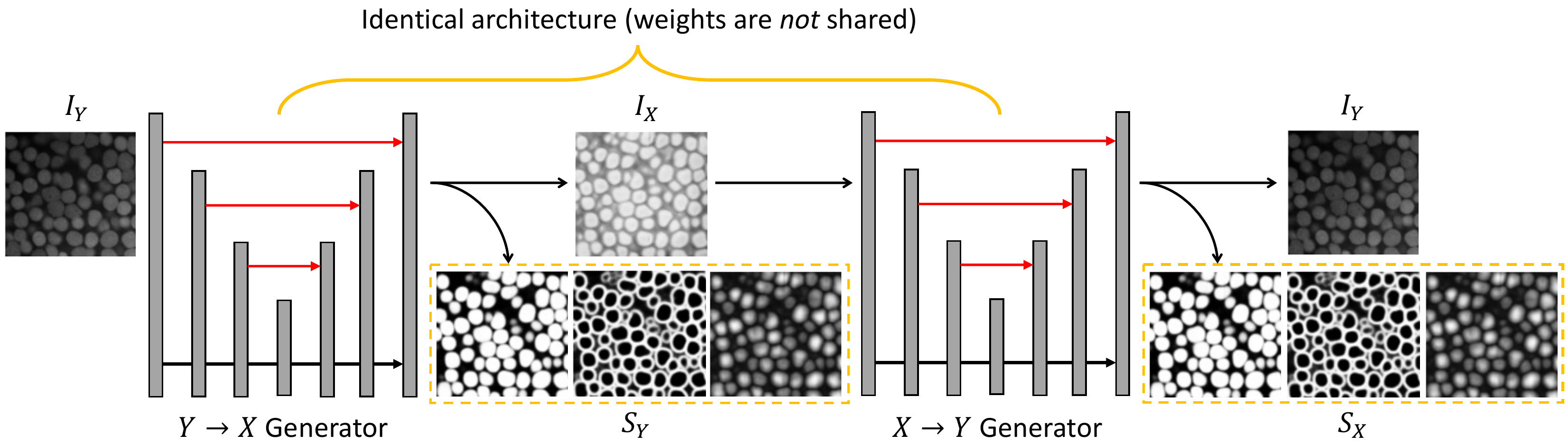}
\caption{
Architecture details of CySGAN. Given an image sampled from $I_Y$, the {\em backward} generator predicts both the transferred image in $I_X$ and the segmentation representations. Then the {\em forward} generator takes only the translated image as input and predicts both the reconstructed image and segmentation representations. The two generators have exactly the same architecture, but the weights are {\em not} shared as they translate images in different domains. 
}
\label{fig:supp_arch}
\end{figure}
\begin{figure}[th]
\centering
\includegraphics[width=\columnwidth]{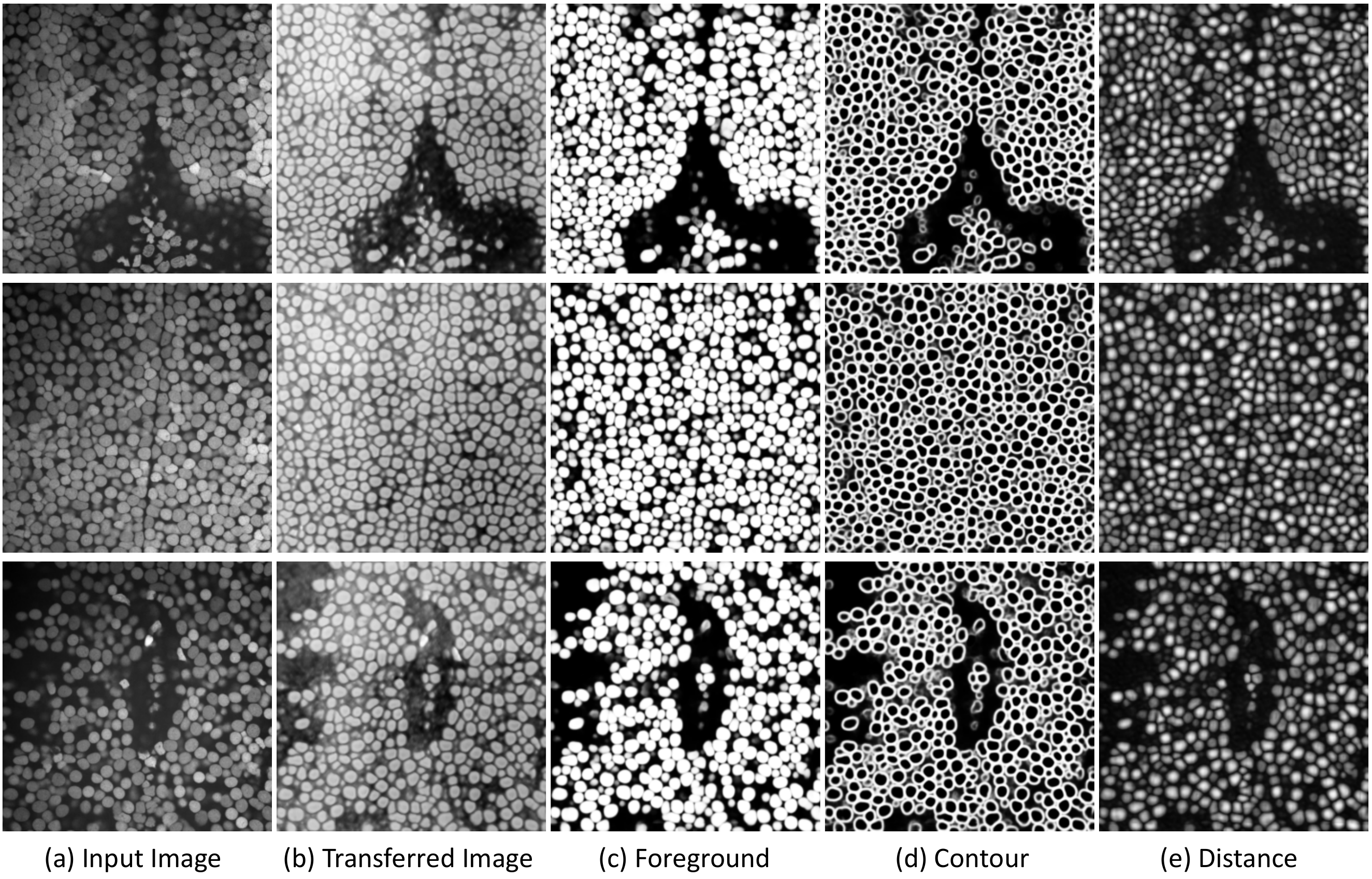}
\caption{
Qualitative results of CySGAN. We show multiple slices of {\bf (a)} input NucExM images, as well as {\bf (b)} transferred images, {\bf (c)} predicted binary foreground masks (B), {\bf (d)} predicted instance contour maps (C) and {\bf (e)} predicted distance transform maps (D) of our proposed CySGAN model. 
}\label{fig:supp_pred}
\end{figure}

\begin{figure}[th]
\centering
\includegraphics[width=\columnwidth]{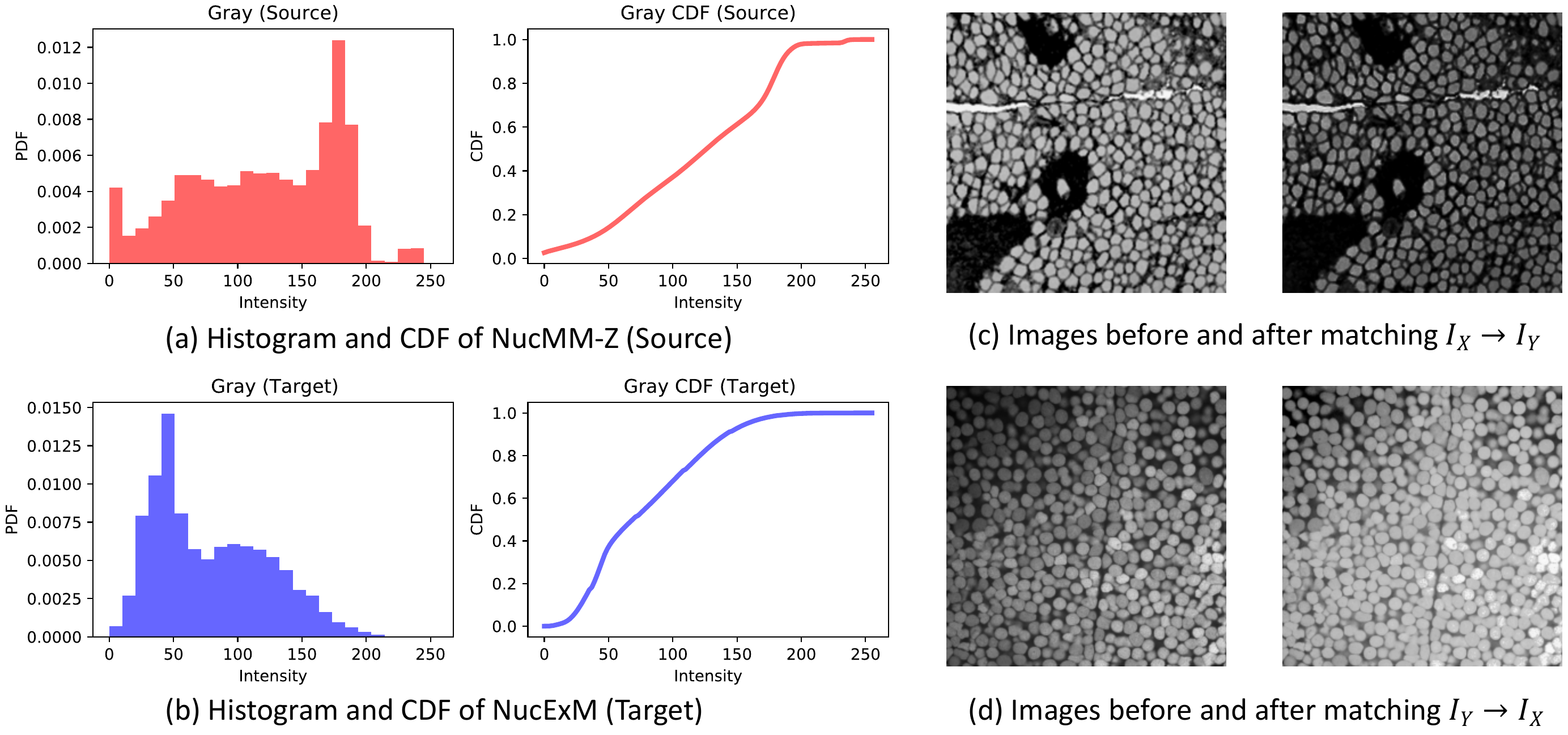}
\caption{
Histogram matching between EM and ExM images. We show the histograms and cumulative distribution functions (CDFs) of {\bf (a)} electron microscopy (EM) and {\bf (b)} expansion microscopy (ExM) images. The effect of histogram matching is shown in {\bf (c)} and {\bf (d)} for both matching directions.
}\label{fig:supp_hist}
\end{figure}
\begin{figure}[th]
\centering
\includegraphics[width=\columnwidth]{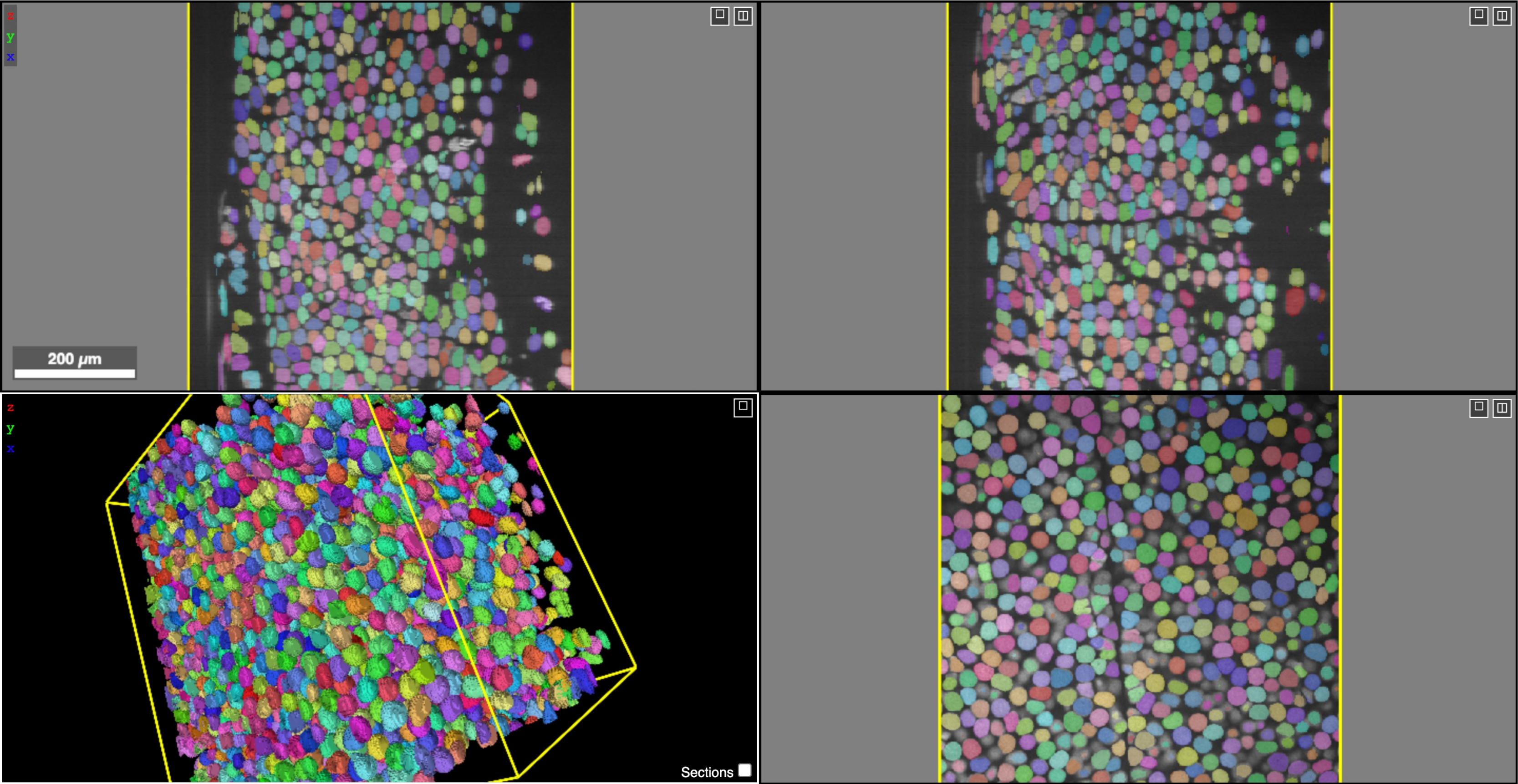}
\caption{
Multi-view visualization and 3D meshes of NucExM. We show the composite views of microscopy images and instance masks of $V_1$ in the NucExM dataset, as well as the 3D meshes of nuclei. We generated the visualizations using the Neuroglancer (\url{https://github.com/google/neuroglancer}).
}\label{fig:np}
\end{figure}

% appendix, comment it for the conference submission
% \input{sections/appendix}
\end{document}